\documentclass[sn-basic,iicol]{sn-jnl}% Basic Springer Nature Reference Style/Chemistry Reference Style with double column layout

\usepackage{natbib}	% to format citation better
\usepackage{tabularx}	% to achieve a table
\theoremstyle{thmstyleone}%
\theoremstyle{thmstyletwo}%

\theoremstyle{thmstylethree}%

\begin{document}

\title[Article Title]{Automated scholarly paper review: Concepts, technologies, and challenges}


\author[1,3]{\fnm{Jialiang} \sur{Lin}}\email{me@linjialiang.net}

\author[1,3]{\fnm{Jiaxin} \sur{Song}}\email{songjiaxin@stu.xmu.edu.cn}

\author[2]{\fnm{Zhangping} \sur{Zhou}}\email{zhouzp@xmu.edu.cn}

\author[1,3]{\fnm{Yidong} \sur{Chen}}\email{ydchen@xmu.edu.cn}

\author*[1,3]{\fnm{Xiaodong} \sur{Shi}}\email{mandel@xmu.edu.cn}

\affil[1]{\orgdiv{School of Informatics}, \orgname{Xiamen University}, \orgaddress{\city{Xiamen}, \country{China}}}

\affil[2]{\orgdiv{College of Foreign Languages and Cultures}, \orgname{Xiamen University}, \orgaddress{\city{Xiamen}, \country{China}}}

\affil[3]{\orgdiv{Key Laboratory of Digital Protection and Intelligent Processing of Intangible Cultural Heritage of Fujian and Taiwan}, \orgname{Ministry of Culture and Tourism}, \orgaddress{\country{China}}}


\abstract{Peer review is a widely accepted mechanism for research evaluation, playing a pivotal role in academic publishing. However, criticisms have long been leveled at this mechanism, mostly because of its poor efficiency and low reproducibility. Recent years have seen the application of artificial intelligence (AI) in assisting the peer review process. Nonetheless, with the involvement of humans, such limitations remain inevitable. In this paper, we propose the concept and pipeline of automated scholarly paper review (ASPR) and review the relevant literature and technologies of achieving a full-scale computerized review process. On the basis of the review and discussion, we conclude that there is already corresponding research and preliminary implementation at each stage of ASPR. We further look into the challenges in ASPR with the existing technologies. The major difficulties lie in inadequate data, imperfect document parsing and representation, defective human--computer interaction, and flawed deep logical reasoning. Moreover, we point out the future directions and discuss the possible moral and ethical issues of ASPR. In the foreseeable future, ASPR and peer review will coexist in a reinforcing manner before ASPR is able to fully undertake the reviewing workload from humans.}

\keywords{Automated scholarly paper review, Peer review, Academic publishing, Natural language processing, Artificial intelligence}



\maketitle

\section{Introduction}
\label{sec:intro}

Peer review, also known as refereeing, has been defined as ``the process of subjecting an author's manuscript to the scrutiny of others who are experts in the same field, prior to publication in a journal''. ``This review process varies from journal to journal but it typically consists of two or three reviewers reporting back to a journal editor who takes the final decision''.~\citep{ware-stm-2015} In the current journal publication process, peer review has become a common practice that plays an extremely important role. It helps editors decide whether academic work can be accepted for publishing in journals. Apart from journal papers, some publications including conference papers and research proposals are also subject to peer review. They generally go through a peer review process similar to that of journal papers. In this study, we mainly focus on journal paper review.

In peer review, whether or not a manuscript can be accepted for publication depends on the review both by journal editors and experts in a certain field. The peer review panel makes a comprehensive assessment on the manuscript's originality, soundness, clarity, and significance to maintain a consistently high standard in scholarly publication. Therefore, peer review is regarded as the ``gatekeeper, the final arbiter of what is valued in academia''~\citep{marsh-improving-2008}. Through peer review, flaws in manuscripts can be found and questionable manuscripts can be intercepted before publication. While at the same time, peer review also serves as a coach in academia as it gives valuable feedback to authors for revision. This helps authors improve the quality of their manuscripts.

Although peer review is widely considered as the norm and its necessity is recognized by the majority of researchers, criticism from both academic and publishing circles has long been leveled at the peer review system~\citep{smith-peer-2006,brezis-arbitrariness-2020}. On the whole, it is subject to criticism surrounding the following issues.

\begin{itemize}

\item \textbf{Poor efficiency.} According to data from~\citet{huisman-duration-2017}, the average duration of peer review, not including the revision, for a manuscript, is 17 weeks. To the author of the manuscript, this lengthy process is a profligate of academic time. In some cases, the research may even become outdated during the inappropriately long process. To researchers who dedicate to review scholarly papers, with the ever-growing number of manuscripts, this reviewing work is an increasingly onerous burden. At the same time, reviewers usually receive few explicit rewards for their hard work. Even though some platforms such as Web of Science\footnote{\url{https://webofscience.help.clarivate.com/en-us/Content/peer-review-in-wos-researcher-profile.html}} and ORCID\footnote{\url{https://info.orcid.org/documentation/workflows/peer-review-workflow/}} attempt to record and credit peer review, there still lack enough incentives for researchers to do the reviewing and rewards to justify such dedication. The time spent on reviewing could be spent alternatively on something more productive, like original research. That being the case, editors are scrambling to find qualified reviewers to even start the reviewing, making the originally time-consuming process even more lengthy. Such a long-drawn process is for sure an unreasonable discouragement for researchers from academic publishing.

\item \textbf{Low reproducibility.} Reproducibility is an important evaluation dimension of science. However, since the subject of peer review is human, different people will have different evaluations of the same thing, and reviewers are no exception. This leads to a certain extent to the uncontrollable situation of peer review. Many studies have shown that peer review is highly random and has low reproducibility. \citet{peters-peer-1982} selected 12 papers written by researchers from leading psychology departments in the U.S.\ and published in 12 different prestigious American psychology journals, which are known for their rigorous peer review process. The papers were resubmitted to the original journals in which they were published under the name of fabricated authors from different institutions. Of these submissions, 3 papers were identified as resubmissions, whereas the remaining 9 papers were sent for external review. Among these 9 papers, 8 were rejected mainly due to ``serious methodological flaws''. Corinna Cortes and Neil Lawrence suggested the arbitrariness in peer review, proving in an experiment that one-half to two-thirds of NIPS 2014\footnote{Conference on Neural Information Processing Systems with its abbreviation later changed to NeurIPS in 2018.} accepted manuscripts might have been rejected if another independent round of review was conducted~\citep{langford-arbitrariness-2015}.

\item \textbf{Limited knowledge.} Due to physical and cognitive limitations, an individual human being is certain to have knowledge limits. Peer review relies on the professional judgments of experts in a certain field. However, expert reviewers still have their knowledge scope within the field and hence the limits. Knowledge is becoming more specific with more emerging subdivisions within a knowledge field. What complicate the reviewing more are the ever-accelerating knowledge explosion and the growing trend of interdisciplinary research. It is more common that the content of the manuscripts goes beyond the scope of knowledge of experts from the same field but not the same subdivisions. The lack of knowledge will lead to the neglect of problems. A flawed manuscript might be wrongfully accepted for publication. This will bring harm to the prestige of the publisher and worst still the development of science.

\item \textbf{Mismatched reviewers.} To any editor, finding suitable reviewers for a given manuscript is no easy task. Corresponding authors of previously published papers in the related domains from the same journal are often chosen as suitable reviewers. This practice, however, does not necessarily make sense because usually under the same domain, there are many different subfields. Experts in one subfield might not be knowledgeable enough to review and comment on works from another subfield. There have been some solutions proposed to tackle this issue~\citep{anjum-pare-2019,pradhan-proactive-2021}, but the recommended researchers are not always available when invited for the reviewing, leaving the solution not very practical.

\item \textbf{Conflicted interests.} Although the academic community has a rigorous system and strict academic regulations, conflicts of interest do occur from time to time. In the peer review system, reviewers are provided with extensive power and there are ways to abuse it. Reviewers could use missing citations as a rhetorical excuse to require authors to include the reviewers' own works in the references while their works might not be strongly related to the current paper. Severer examples can also be found as there is a loophole in this system that ideas can be stolen. A reviewer of \textit{Annals of Internal Medicine} plagiarized a manuscript he reviewed and published it in another journal with data fraud at the same time~\citep{laine-annals-2017}. Plagiarism committed in peer review is an extreme example of reviewers abusing their power, but such examples do prove the possible damage the flawed system of peer review can cause.

\end{itemize}

Academic publishers use peer review to ensure the quality and integrity of science in publication. However, the problems above lead to flawed evaluation and hinder the development of science, which are running counter to the purpose of peer review. To young scientists, the defective system is especially harmful to their academic careers. The difficulty in publication can dampen their enthusiasm for doing research. Peer review is supposed to improve science, but it may now do exactly the opposite.

The term artificial intelligence (AI) first appeared in public view in 1956 when John McCarthy (1927--2011), Marvin Minsky (1927--2016), and some other experts in the related fields met at Dartmouth College for the Dartmouth Summer Research Project on Artificial Intelligence. Many well-known scholars have defined AI. \citet{collins-artificial-2021} has collected 28 definitions of AI from primary studies. Here, we choose the definition given by Arun Rai, an expert in the area of information systems. The definition is detailed and clear, compatible with the contemporary view of AI, and has gained wide recognition in the academic community upon its presentation. ``AI is typically defined as the ability of a machine to perform cognitive functions that we associate with human minds, such as perceiving, reasoning, learning, interacting with the environment, problem solving, decision-making, and even demonstrating creativity''.~\citep{rai-next-2019} AI ``is concerned with not just understanding but also building intelligent entities---machines that can compute how to act effectively and safely in a wide variety of novel situations''~\citep{russell-artificial-2021}.

With the development of AI, computers have been able to defeat the best human players in many fields, including Go game~\citep{silver-mastering-2017}, Texas hold'em poker~\citep{brown-superhuman-2019}, and esport Dota 2~\citep{openai-dota-2019}. The latest generation AI is also competent to predict protein structure~\citep{jumper-highly-2021}, perform quantitative reasoning~\citep{lewkowycz-solving-2022}, and participate in programming competitions~\citep{li-competition-2022}.

Naturally a question arises: is AI already developed well enough to review scholarly papers independently? To answer this question, we propose and enunciate the concept of automated scholarly paper review (ASPR). Then we review related technologies required at each stage of the ASPR pipeline. Some of them have already been incorporated into the peer review process. Some are developed for other purposes and can be employed to achieve ASPR. Further, we discuss the challenges that these technologies face when used for ASPR and look into their future development. In this paper, we mainly study English-related technologies. With some modifications, these technologies can be used to process multilingual manuscripts in ASPR\@.

\section{Concepts}

\textbf{Automated scholarly paper review (ASPR) is defined as a process that computers or other intelligent machines independently review the content of a scholarly paper and automatically generate a review report which typically consists of a summary, review comments, scores, and a decision, etc. for a specific publication venue.}

Firstly, it is plainly specified in this concept that the review subject is and only is ``computers or other intelligent machines'' (hereinafter referred to as computers), not humans. The entire process is performed independently and automatically by computers without any human involvement. This breaks away from the traditional concept of using computers as an assistant for reviewing in previous studies. Computers are designed to complete the whole review process all on their own, fulfilling the roles of both editors and reviewers in an absolute sense. The avoidance of phrases such as ``using computers'' or ``utilizing computers'' in the wording of the concept indicates that computers are not tools, but an existence with the possession of humans status in the process.

Secondly, it is emphasized in this concept that the review subject of ``computers or other intelligent machines'' are ``intelligent''. As defined by~\citet{russell-artificial-2021}, an intelligent machine should perform optimally in the best possible manner. An intelligent machine is not expected to actually ``comprehend'' the manuscript submitted for review, which would be unnecessary. As per the conventional notion of AI, a machine that exhibits ``rational'' behavior is to be attributed with intelligence, irrespective of the underlying mechanism and rationale to achieve such intelligence.

Thirdly, it is stated in the concept that the review object is ``the content of a scholarly paper''. This entails two restrictions on the scope of review object. First, ASPR pertains to the review of scholarly papers (hereinafter referred to as papers), which follow a set of rigorous norms and conventions, and are evaluated through specific procedures. Although other types of research documents that share similar features of papers, such as research proposals, can also be evaluated with the technology similarly used for papers, they are not included in the definition of this concept for the sake of academic rigor. Second, the ``content'' of papers to be reviewed is restricted to the title, abstract, main body, figures, tables, footnotes, references, and even external sources referred to via hyperlinks in the paper. ASPR does not take into account any information about the authors, such as their affiliations, positions, or reputation. It can be construed as a form of review with concealed author identity. Also, it should be noted that there are previous studies~\citep{ruan-alternative-2018} that assess papers based on post-publication information, including citation counts, downloads, Altmetric Attention Scores, etc. Such information is typically collected and analyzed after the publication and is not available during the peer review process. The process of ASPR aligns with the reality of peer review, whereby the information used by computers for manuscript reviewing is consistent with that of the process of peer review. Only the aforementioned restricted aspects of content and all human knowledge in the current context are taken into consideration. ASPR does not assess manuscripts over additional post-publication information.

Fourthly, the concept describes the review scenario as ``a specific publication venue''. This implies that the review circumstances, including criteria, requirements, and results of papers vary with different publication venues. This is also the case for peer review where the same manuscript can be rejected by Journal A and accepted without any revision later by Journal B. Various factors come into play for the different decisions, including subjective ones like the preferences of editors-in-chief and the criteria of different reviewers. Apart from the subjective factors, there are also objective factors from the journals influencing the acceptance of manuscripts. Different journals have their own specific aims and scopes, review standards, and publication requirements. For instance, the criteria of high impact factor journals like \textit{Nature} and \textit{Science} are certain to differ from those of some predatory journals with lower impact factors. ASPR employs applicable criteria, i.e.,~the criteria of a specific publication venue that the manuscript is submitted to.

Fifthly, the concept proposes that the output of ASPR is a ``review report'' that includes but not limits to ``a summary, review comments, scores, and a decision'', which are also the four regular components of a review report in peer review. In light of the fact that ASPR is designed to be analogous to peer review, the results of ASPR need to be presented with these four components as well. It should be noted that the components are to be generated based on the criteria of a ``specific publication venue'' as described above.

Last but not least, it is established in this concept that ASPR is defined by the core term of ``process''. In other words, ASPR covers a series of stages from manuscript receiving, parsing, screening, reviewing, commenting, and scoring to final decision making. Everything included in this process falls within the scope of ASPR research, which may also include moral and ethical issues, etc. Some of the specific tasks, such as scoring, acceptance prediction, and review comment generation, are only specific technologies used to facilitate ASPR rather than being the representatives of ASPR itself.

\begin{figure}[htb]
\centering
\includegraphics[trim={1.3in 1in 1.4in 1in},clip,width=0.47\textwidth]{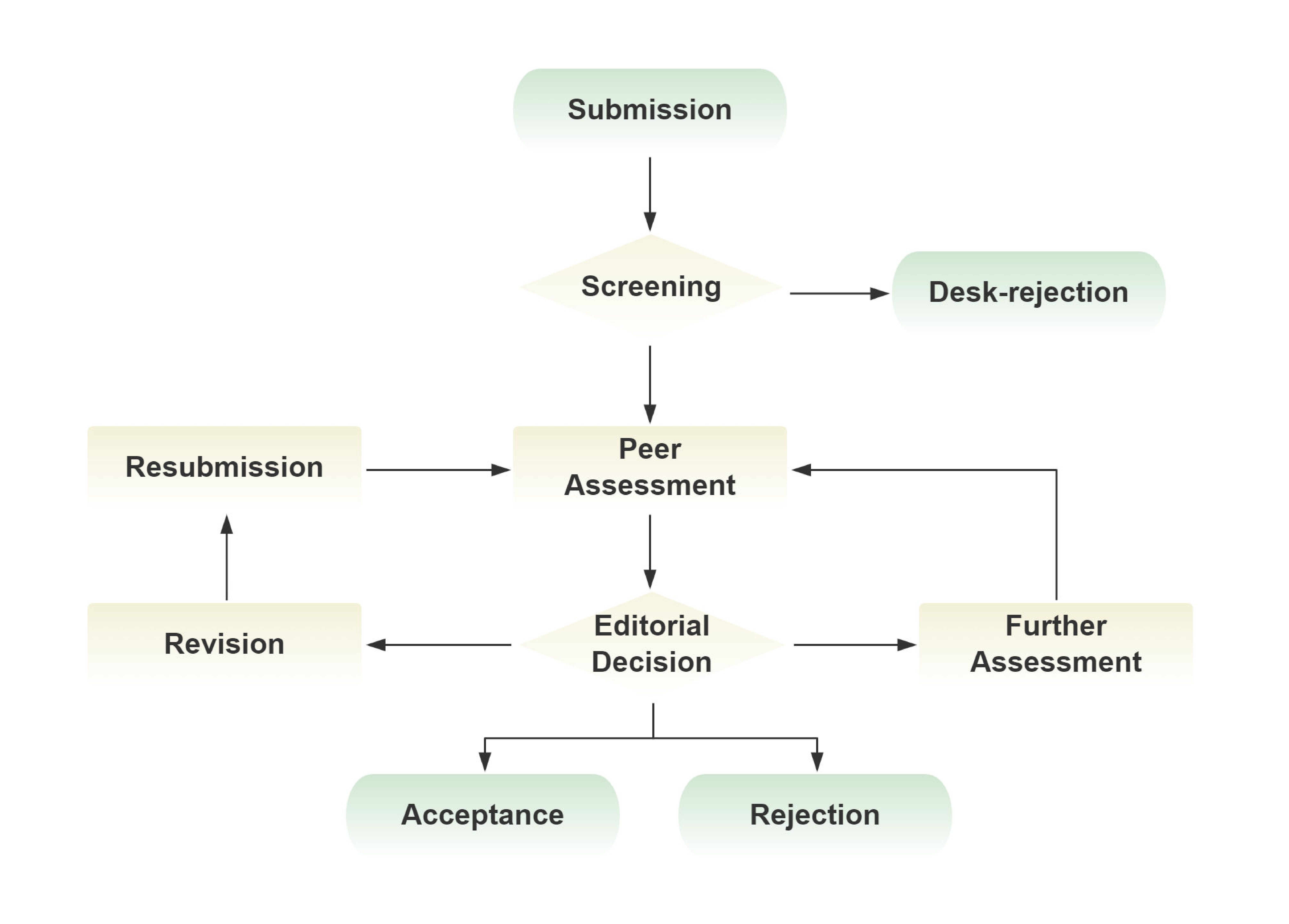}
\caption{Peer review pipeline.}
\label{fig:peer-review-pipeline}
\begin{tablenotes}
	\item \footnotesize{Note: This pipeline is based on the peer review process of several prestigious publishers, including Elsevier (\url{https://www.youtube.com/watch?v=sRgmTtVKj9I}), Springer Nature (\url{https://www.biomedcentral.com/getpublished/peer-review-process}), Wiley (\url{https://authorservices.wiley.com/Reviewers/journal-reviewers/what-is-peer-review/the-peer-review-process.html}), etc.}
\end{tablenotes}
\end{figure}

\begin{figure*}[htb]
\centering
\includegraphics[trim={0 0.8in 0 0.8in},clip,width=0.76\textwidth]{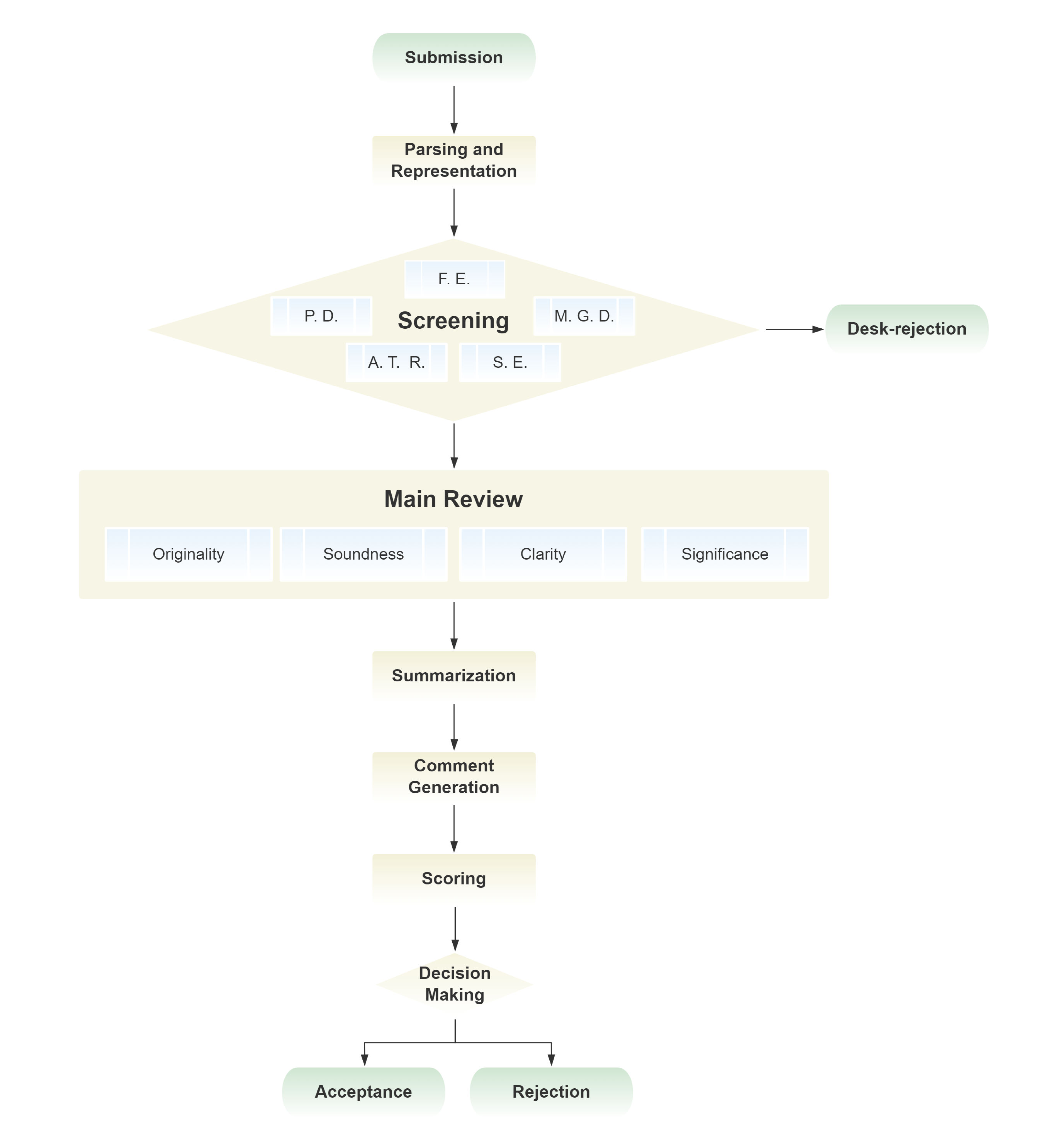}
\caption{Automated scholarly paper review pipeline.}
\label{fig:aspr-pipeline}
\begin{tablenotes}
	\item \footnotesize{Note: In Screening, F. E. is short for Format Examination, P. D. for Plagiarism Detection, M. G. D. for Machine Generation Detection, A. T. R. for Article Type Recognition, and S. E. for Scope Evaluation.}
\end{tablenotes}
\end{figure*}

Peer review is a process with multiple parties involved, typically including an editor-in-chief, a review editor, at least two reviewers, and the authors. The overall process of traditional peer review is shown in Fig.~\ref{fig:peer-review-pipeline}. In contrast, ASPR is an all-in-one solution with computers functioning as the editor-in-chief, the review editor, and the reviewers simultaneously to complete the entire review process independently. The pipeline of ASPR is illustrated in Fig.~\ref{fig:aspr-pipeline}.

\section{Research rationale}

Shifting the current workload of peer review from humans to computers comes with the following advantages.

ASPR is capable of overcoming human limitations, both physically and psychologically. Physically, there is a limit to how much information a normal human being can handle. No matter how knowledgeable a person is, one can only grasp part of the total knowledge existing in this world. Psychologically, there is a limit to how objective a person can be. It is almost inhuman to expect a human to remain fair and objective completely and constantly in evaluation. These weaknesses to humans are exactly the strengths of computers. Theoretically, all knowledge in the world can be gathered by computers to create a unified knowledge base, which can then be used for information processing by themselves. Computers themselves are not affected by emotions, hence the non-existence of conflicts of interest in reviewing.

ASPR can tackle the issue of the overwhelmingly increasing volume of submitted manuscripts. According to data from Dimensions,\footnote{\url{https://app.dimensions.ai/discover/publication}} the number of scholarly publications has been increasing constantly from 2011 to 2020, from about 3.7 million to around 6.7 million.\footnote{Data accessed in May 2023.} The number of manuscripts posted to preprint platforms such as arXiv\footnote{\url{https://arxiv.org/}} is also on the rise year by year. As reported by~\citet{lin-how-2020}, from 2008 to 2017, the preprints posted on arXiv in the category of computer science total 141,961, 77.1\% of which are published on peer-reviewed venues. The number of published papers itself is overwhelming enough and to add to this is the similarly enormous amount of those rejected papers which have also gone through the peer review process. To make it trickier, one manuscript generally needs at least two reviewers to counter subjectivity. All these circumstances together pose an overwhelming workload to the peer review system. To humans, these figures might look daunting, but to computers, they are not, because this is what computers are invented for. With proper architectural design and hardware support, computers can reach enormous computing power to review manuscripts in a very short period of time.

ASPR helps authors improve their manuscripts in a highly efficient manner. In the traditional peer review system, authors usually need to wait several months before the editorial decision is made. If a manuscript is returned for revision, it needs to be revised based on reviewers' comments and then be resubmitted for another round of review. This process is really time-consuming. Whereas, ASPR is different. A manuscript submitted to an ASPR system for reviewing can have feedback given immediately. Such instant response help authors save more time to make improvements to their manuscripts. Also, apart from being used by the publishers for manuscript reviewing before publication, ASPR can serve as an academic writing checker provided to individual authors in their writing.

ASPR, as a highly integrated and demanding task, is beneficial to many related technologies. The introduction of a new integrated task often relies on related technologies and in turn this new task can drive the development of these related technologies. This is the case for ASPR. The realization of ASPR can only be achieved with the most advanced technologies in several different fields. Examples include document parsing, document representation, text classification, information retrieval, text generation, etc. The development of these technologies will also drive the emergence of more relevant datasets, creating a virtuous circle. At the same time, ASPR will in turn bring more studies into these fields. With the proposal of ASPR, we aim to bring overall progress to science and technology in different fields as a whole.

ASPR can significantly cut down the high cost of time and money in the current peer review. The research conducted by~\citet{houghton-open-2009} studied alternative models for academic publishing from the perspective of economic factors. The cost of time and money in the peer review process was quantified in their research and it was estimated that the average cost of a paper was about 1,440 euros. This research is based on the price level in 2007. Taking into consideration of the price level at present, the cost of each peer-reviewed paper must have increased a lot more. From this study, we can see that the overall cost of peer review is excessively high. ASPR is a highly economical alternative to the costly peer review process both in terms of time and money. With the application of ASPR, human reviewers can be freed from the increasingly heavy workload to devote more time and energy to their own research; the publication cost can be reduced to a great extent both for the publishers and the authors.

\section{Related work and datasets}

To our best knowledge, there is no previous paper that systematically and comprehensively investigates the whole process of achieving ASPR. Efforts have been made in previous studies to use computers in assisting peer review and improving the editorial work, but their focus only lies on computer assistance leaving a blank in automating the whole process of peer review. \citet{price-computational-2017} used computers to match manuscripts with suitable reviewers, assemble balanced peer review panels, and calibrate review scores. \citet{mrowinski-artificial-2017} used an evolutionary algorithm to optimize the editorial workflow, greatly reducing the peer review time by 30\% without relying on more reviewers. \citet{heaven-ai-2018} demonstrated various AI tools that can help publishers improve the peer review process with computational support in choosing reviewers, validating data, summarizing key findings, etc. \citet{checco-ai-2021} explored the strengths, the weaknesses, and the uncertainty in using AI to assist the reviewing of research results. Currently, most researchers hold the view that AI is an assistant to editors and reviewers but not a replacement. Table~\ref{tab:existing-tools} shows the assisting tools for peer review.

\begin{table*}
 \caption{Existing tools of assisting peer review process.}
 \label{tab:existing-tools}
	\centering
	\small
	\begin{tabularx}{\textwidth}{lclX}
  \hline\noalign{\smallskip}
  \textbf{Name}  &  \textbf{Year}  &  \textbf{Official website}  &  \textbf{Applications}  \\
  \hline\noalign{\smallskip}

  \textbf{CrossCheck} & 2008 & \url{https://crosscheck.ithenticate.com/} & Plagiarism checking, improper reference detection, writing scoring  \\
  \hline\noalign{\smallskip}

  \textbf{UNSILO} & 2012 & \url{https://unsilo.ai/} & Key concept extraction, linguistic quality assessment, journal match reviewing, reviewer finder, manuscript screening  \\
  \hline\noalign{\smallskip}

  \textbf{Recite}  & 2014  & \url{https://reciteworks.com/} & Citation match checking, reference format checking  \\
  \hline\noalign{\smallskip}

  \textbf{Statcheck} & 2014 & \url{http://statcheck.io/} & Statistical data extraction, \textit{P} values recomputation and consistency checking  \\
  \hline\noalign{\smallskip}

  \textbf{Penelope} & 2016 & \url{https://www.penelope.ai/} & Matching journal request detection, missing reference checking, figure citation checking, standard format checking  \\
  \hline\noalign{\smallskip}

  \textbf{StatReviewer} & 2016 & \url{http://statreviewer.com/} & Numerical error checking, statistical tests, integrity and quality checking, methodological reporting  \\
  \hline\noalign{\smallskip}

  \textbf{Scholarcy}  & 2018  & \url{https://www.scholarcy.com/} & Paper summarization, reference extraction, table and figure extraction, chapter segmentation and highlighting  \\
  \hline\noalign{\smallskip}

  \textbf{SciScore} & 2019 & \url{https://sciscore.com/} & Key resource analyzing, NIH, MDAR, and ARRIVE report generation, journal reproducibility tracking with RTI  \\
  \hline\noalign{\smallskip}

  \textbf{ReviewAdvisor} & 2021 & \url{http://review.nlpedia.ai/} & Review comment generation from eight aspects: originality, substance, replicability, clarity, etc. \\
  \hline\noalign{\smallskip}

\end{tabularx}
\begin{tablenotes}
  \item \footnotesize{Note: Links accessed in May 2023 and the link to ReviewAdvisor was not working at the time. According to the README file on its GitHub repository, the link is ``currently offline due to computational consideration'' and its authors ``will re-start it once the computational resource is sufficient''. For more information please visit its GitHub repository: \url{https://github.com/neulab/ReviewAdvisor}.}
  \end{tablenotes}
\end{table*}

This research falls under the general domain of automated evaluation of text quality. A highly similar task to ASPR is automated essay scoring (AES)~\citep{ke-automated-2019,ramesh-automated-2022}. In the 1960s, \citet{page-imminence-1966} released the Project Essay Grade, which later becomes the representative of the AES system. In this system, the scoring is based on basic linguistic features. Another well-known system is Intelligent Essay Assessor~\citep{foltz-intelligent-1999}. This system grades an essay by making a comparison between this given essay and the outstanding essays stored in the system using Latent Semantic Analysis (LSA)~\citep{deerwester-indexing-1990}. Educational Testing Service (ETS) developed e-rater~\citep{attali-automated-2004} for the scoring in the Test of English as a Foreign Language (TOEFL). This tool is based on continuously updated natural language processing (NLP) techniques and is widely accepted by the English language qualification community. An essay is defined as ``a short piece of writing on a particular subject, especially one done by students as part of the work for a course'' in the \textit{Cambridge Dictionary}.\footnote{\url{https://dictionary.cambridge.org/dictionary/english/essay}} Compared to an essay, a paper is normally longer in length and more complex in structure. The most different part is that papers, or at least those quality ones, should present new ideas or findings. Therefore, the standards in essay evaluation and paper evaluation should have different focuses. But the techniques used in AES can also work for ASPR. Other applications of automated evaluation of text quality include content assessments on Wikipedia articles~\citep{marresetaylor-edit-2019}, evaluation of TED talks~\citep{tanveer-causality-2019}, and prediction of postgraduate program admission based on the Statement of Purpose~\citep{kanojia-is-2017}. All these tasks like AES have related techniques that can be used to achieve ASPR\@.

To AI research, data is especially crucial. The training and evaluation of research models cannot go without data. The realization of ASPR needs adequate annotation data, such as review comments, review scores, editorial decisions, etc.

NeurIPS is a prestigious conference in machine learning. This conference offers to the public their conference papers along with review comments, meta-reviews, and discussion records between the reviewers and the authors.\footnote{\url{https://neurips.cc/Conferences/2022/CallForPapers}} Apart from NeurIPS, reviewing information can also be found in some journals, such as eLife\footnote{\url{https://reviewer.elifesciences.org/author-guide/editorial-process}} and PeerJ,\footnote{\url{https://peerj.com/benefits/review-history-and-peer-review/}} and also platforms like OpenReview\footnote{\url{https://openreview.net/about}} and F1000Research.\footnote{\url{https://f1000research.com/about}} However, the problem is that most rejected manuscripts are not open to the public, which will lead to data imbalance between accepted and rejected manuscripts, thus affecting the effectiveness of models developed based on the data.

In addition to using public raw data, some researchers have built structured datasets for research purposes. We select some of the datasets and their corresponding applications to be presented in Table~\ref{tab:dataset-list} with their properties given respectively. These datasets of various domains can be used at different stages of automated review, paving the foundation for the achievement of ASPR. In addition, these datasets also enable aspect performance comparisons of tasks within ASPR\@.

\begin{table*}
 \caption{ASPR related datasets.}
 \label{tab:dataset-list}
	\centering
	\small
	\resizebox{0.82\paperwidth}{!}{
	\begin{tabularx}{\paperwidth}{llXX}
  \hline\noalign{\smallskip}
  \textbf{Application}  &  \textbf{Name}  &  \textbf{Scale} &  \textbf{Contents} \\

  \hline\noalign{\smallskip}
  \multirow{2}*{\textbf{Citation analysis}}
  & S2ORC~\citep{lo-s2orc-2020} & 8.1M papers with GROBID full text, 1.5M papers with parsed citations & Metadata, full text, in-text citations \\
  & unarXive~\citep{saier-unarxive-2020} & 1M papers, 29.2M citations and corresponding contexts & Metadata, full text, in-text citations \\

  \hline\noalign{\smallskip}
  \multirow{3}*{\textbf{Language assessment}}
  & LEDAT~\citep{daudaravicius-language-2014} & 4k papers & Source text, text edited by professional editors \\
  & Sentence-level revisions~\citep{tan-corpus-2014} & 108k sentence pairs  & Sentences comparison pairs of different versions, strength annotations \\
  & TOEFL-Spell~\citep{flor-benchmark-2019} & 6k spelling errors based on 883 essays & Annotated spelling errors in contexts \\

  \hline\noalign{\smallskip}
  \textbf{Novelty detection}
  & TAP-DLND 1.0~\citep{ghosal-tap-2018} & 6k newspapers & Document with non-novel or novel annotation \\

  \hline\noalign{\smallskip}
  \multirow{6}*{\textbf{Reviewing}}
  & PeerRead~\citep{kang-dataset-2018} & 14.7k papers, 10.7k textual peer reviews & Full text, decisions, scores, reviews \\
  & ACL-18 Numerical~\citep{gao-does-2019} & 4k review comments, 1.2k author responses & Decisions, scores before and after rebuttal, review comments, response text \\
  & CiteTracked~\citep{plank-citetracked-2019} & 3k papers & Metadata, review comments, citations \\
  & Interspeech 2019 Submission~\citep{stappen-uncertainty-2020} & 2k submissions, 6k review comments & Metadata, decisions, scores, review comments \\
  & COMPARE~\citep{singh-compare-2021} & 39 papers with 117 review comments covering 1.8k sentences & Decisions, categories, comparison discussion sentences \\
  & ASAP-Review~\citep{yuan-can-2022} & 9k papers & Full text, decisions, review comments texts with aspect annotations \\

  \hline\noalign{\smallskip}
  \multirow{2}*{\textbf{Scoring}}
  & AAPR~\citep{yang-automatic-2018} & 19k papers & Full text, decisions \\
  & ACL-BiblioMetry~\citep{dongen-schubert-2020} & 30k papers & Metadata, full text, citation scores \\

  \hline\noalign{\smallskip}
  \textbf{SOTA comparison}
  & TDMSci~\citep{hou-tdmsci-2021} & 2k sentences with around 3k TDM entities & Task, dataset, metric (TDM) entities  \\

  \hline\noalign{\smallskip}
  \multirow{4}*{\textbf{Summarization}}
  & ScisummNet~\citep{yasunaga-scisummnet-2019} & 1k high cited papers with corresponding citation sentences & Full text, citation information, manual summaries \\
  & TalkSumm~\citep{lev-talksumm-2019} & 1.7k papers with video talks & Full text, corresponding talk transcripts \\
  & SciTDLR~\citep{cachola-tldr-2020} & 5.4k TLDRs over 3.2k papers & Full text, author-written and expert-derived TLDRs \\
  & FacetSum~\citep{meng-bringing-2021} & 60k papers & Full text, faceted summarizations target on specific sections \\

  \hline\noalign{\smallskip}
	\end{tabularx}
	}
\end{table*}

\section{Parsing and representation}
\label{sec:parsing-representation}

In general, when submitting a manuscript to an academic venue, the manuscript is required to be formatted in LaTeX or PDF. In traditional peer review, reviewers first need to use appropriate tools to open files for reading and pass the content to their brains for further processing. Similar to the peer review process, the first step for ASPR is to conduct parsing and representation. In this prerequisite step, different unstructured contents, including texts, figures, tables, mathematical expressions, and related metadata, need to be parsed and represented into proper data form before they can be processed by computers.

When it comes to LaTeX files, parsing is relatively easier. This is because LaTeX files are actually well-structured plain text and the files' structure and content are required to be specified with symbols. Parsing manuscripts in the format of PDF is more challenging, because PDF files do not come with structural tags. It takes much more effort to extract data from PDF files while preserving the original layout. Currently, the most commonly used parser for PDF files is GROBID~\citep{lopez-grobid-2009}, which is released in 2009, and has been constantly maintained and updated ever since. When it is first released, GROBID mainly performs the extraction of bibliographical data and is later updated for header extraction and parsing, in-text citation recognition and resolution, full-text extraction and structuring, etc. GROBID is a powerful tool in PDF parsing, nonetheless, it is not very good at figures, tables, and mathematical expressions extracting, which are almost indispensable to papers. Efforts have been made to solve these problems. For figure extraction, \citet{li-figure-2019} presented a tool using layout information to extract figures and their corresponding captions. This tool is available online.\footnote{\url{https://www.eecis.udel.edu/\%7Ecompbio/PDFigCapX}} For table extraction, \citet{zheng-global-2021} proposed Global Table Extractor (GTE) to detect tables and recognize cell structure jointly based on visual context. For mathematical expressions, \citet{wang-extraction-2019} introduced an unsupervised font analysis-based method to extract them from PDF files.

After parsing the submitted files, the next step in ASPR is to represent the structured data in suitable forms. For text, the common representation method is feature vector representation. In 2013, \citet{mikolov-efficient-2013} published the technique word2vec, which converts words into vectors efficiently. There are two methods for learning representations of words in word2vec: the continuous bag-of-words (CBOW) model and the continuous Skip-gram model. CBOW predicts words based on the surrounding context, while in the continuous Skip-gram, the given words are used to predict the surrounding context words. Despite being a great technique of NLP, word2vec is not flexible enough to process polysemous words, i.e.\ words with multiple meanings, as it only generates a single embedding representation for each word or phrase. To produce more semantic-rich word embeddings, \citet{peters-deep-2018} proposed a bidirectional language model ELMo, which was pre-trained on a large amount of text data. Unlike word2vec, when computing word representation for given words, this model takes into account the context, either at sentence or paragraph level, for each occurrence of the words. This way words are allowed to have separate embeddings for different meanings. Similar to ELMo, BERT~\citep{devlin-bert-2019} is also a pre-trained model, but it uses Transformer~\citep{vaswani-attention-2017} to generate representations for words. BERT is then used as the base for many later-come pre-trained models~\citep{qiu-pre-2020}. This makes a significant change in the representation of words, sentences, paragraphs, and documents and ushers in a new era for NLP\@.

For image representation, Convolutional Neural Network (CNN)~\citep{lecun-backpropagation-1989} is the classic and commonly used method. AlexNet is a typical deep architecture with 8 layers. With its much better performance, AlexNet proves that larger and deeper neural networks can better extract image features~\citep{alex-imagenet-2012}. However, as the number of layers increases, the risk of degradation problem also goes up. ResNet is designed by~\citet{he-deep-2016} for this problem and it is proved to be an effective solution. ResNet and its variants are popularly used for image feature extraction. Recently, inspired by the successful application of Transformer in NLP, \citet{dosovitskiy-image-2021} used Transformer directly and achieved outstanding performance.

For table representation, the traditional approach of vectorization was mainly employed in the early stage. \citet{gentile-entity-2017} trained table embedding with clustering to facilitate data blocking for record linkage within tables. \citet{ghasemi-tabvec-2018} presented TabVec, an unsupervised table embedding method used for table category classification. \citet{zhang-table2vec-2019} proposed Table2Vec, a method utilizing table caption, column headings, and cells to train the embedding of words and entities for the tasks of row population, column population, and table retrieval. Recent years have seen the use of pre-training methods for the representation of tables, \citet{dong-table-2022} provided a review of various model architectures, pre-training objectives, and downstream tasks related to the pre-training for table representation.

For mathematical expression representation, \citet{jo-modeling-2021} proposed a method to model the semantics of mathematical expressions and their accompanying text in papers. \citet{mansouri-embedding-2021} presented an embedding model of mathematical expressions and their surrounding text for information retrieval. Similar to text, images, and tables mentioned above, mathematical expressions are also influenced by the pre-trained model in the representation. \citet{peng-mathbert-2021} introduced MathBERT, a pre-trained model trained jointly with mathematical expressions and their corresponding texts. MathBERT captures semantic-level structural features of mathematical expressions through a new pre-training task.

The submitted files are parsed and represented with these various techniques available for further reviewing.

\section{Screening}
\label{sec:screening}

Editorial screening is the initial step of reviewing that decides whether the submitted manuscript should be sent for further reviewing. According to data from \textit{Nature}, most of the manuscripts are desk-rejected without being sent out for external review with a desk-rejection rate of approximately 60\%.\footnote{\url{https://blogs.nature.com/peer-to-peer/2006/12/report_of_natures_peer_review_trial.html}} It is also reported that around 80\% of the manuscripts submitted to \textit{Nature Microbiology} are already rejected after screening~\citep{editorial-interesting-2016}. In order to cope with the overwhelmingly large quantities of submissions, the top AI conference International Joint Conferences on Artificial Intelligence (IJCAI) also designs a similar mechanism called summary-reject.\footnote{\url{https://ijcai20.org/faq/\#summary}} At the screening stage, the detailed content of the manuscripts will not be put under close scrutiny. The manuscripts are mainly checked for their formats, topics, plagiarism, etc. to ensure conformity to the style guidelines and instructions as well as the aims and scope of the journal. Most work can be completed by AI sufficiently and many publishers have been using AI in assisting screening.

\subsection{Format examination}

It is the researchers' responsibility to fulfill the format requirements of the academic venues in their preparation for submission. This is the most basic skill of academic literacy that researchers should acquire. Therefore, format examination is also the basic step in screening to filter out unprofessional submissions with formatting issues. Most format checkers are based on document parsing tools. The PDF format is ubiquitous in submission. Format examination for PDF submissions is mainly targeted at the layers, fonts, length, etc., regardless of the content. One example is PDF eXpress from IEEE.\footnote{\url{https://www.ieee.org/conferences/publishing/pdfexpress.html}} For submissions in unstructured document format, submissions can be parsed to check their differences in formatting with the standard template, for example, the APA Style~\citep{american-publication-2019}. \citet{lu-xml-2014} used the XML format node template tree to detect formatting issues automatically. The model compares the format feature tree of a given paper with that of a template document and checks for differences.

\subsection{Plagiarism detection}

Plagiarism is a serious academic offense, which should be detected at the earliest stage of reviewing as possible. Compared to other techniques involved in ASPR, plagiarism detection has yielded more mature methods, because of its early start in traditional reviewing. Instead of solely relying on the surface string similarity, \citet{osman-improved-2012} used semantic role tagging (SRL) for plagiarism detection. \citet{abdi-plagiarism-2015} combined both semantic and syntactic information to detect external plagiarism. External Plagiarism Detection System (EPDS)~\citep{abdi-linguistic-2017} made further improvements by incorporating SRL technology, semantic, syntactic information, and content word expansion approach into the model. This method can detect multiple kinds of plagiarism, such as paraphrasing, sentence transformation, and word structure changes. \citet{sahi-novel-2017} measured the semantic similarity between papers by computing the similarity of their corresponding topic events, which are the combination of multiple information profiles of the papers. This method can efficiently be used for plagiarism detection. \citet{ahuja-new-2020} designed a more thorough plagiarism detection system, incorporating semantic knowledge like Dice measure, path similarity, and depth estimation measure to compute the resemblance with different weights assigned.

Apart from monolingual plagiarism, cross-language plagiarism is becoming a growing threat to academic integrity. This type of plagiarism occurs when a paper or fragment is translated from another paper or fragment written in a different language without proper citation. The detection of cross-language plagiarism is more complex. \citet{ehsan-candidate-2016} proposed a keyword-based approach to cross-language plagiarism detection. In their method, text is segmented based on different topics for the computing of local similarity. \citet{roostaee-effective-2020} further proposed a method combining concept model with bag-of-words model to make use of both concept and keyword information with dynamic interpolation factor. This method achieved outstanding results in German-English and Spanish-English cross-language plagiarism detection. \citet{gharavi-scalable-2020} introduced a two-step sentence-by-sentence comparison method to detect cross-language plagiarism. Firstly, a comparison is conducted between sentence representations with semantic and syntactic information to measure their similarity for candidate plagiarized documents. Secondly, parameter tunings both online and offline are employed to filter out actual plagiarized documents.

In addition to the plagiarism detection of text, it is equally important to check the images for plagiarism. \citet{eisa-content-2019} identified similar semantics in graphics and detected structural changes with image technology for graphic plagiarism detection. \citet{eisa-text-2020} studied the underlying features of graphics and proposed a more in-depth method, in which the graphic components are analyzed to obtain the meaning of the graphic. \citet{meuschke-analyzing-2021} conducted plagiarism detection by combining the detection of textual content and all other non-textual content for higher efficiency.

\subsection{Machine generation detection}

With the development of natural language generation (NLG), some programs are designed to generate research papers with figures, graphs, and citations included. One famous example is SCIGen.\footnote{\url{https://pdos.csail.mit.edu/archive/scigen/}} There are cases of some researchers using these programs to generate papers for academic publishing. \citet{labbe-duplicate-2013} detected that 85 papers generated by SCIGen have been published. Under these circumstances, identifying machine-generated submissions should become an indispensable part of screening. \citet{amancio-comparing-2015} identified randomly generated nonsense in the form of research papers by creating a complex network through modeling on text. This network achieved at least 89\% accuracy in machine generation detection. One of its contributions lies in that it proves that network features can be used to identify randomly generated meaningless papers. \citet{nguyen-son-identifying-2017} used statistical information to identify machine-generated text based on the differences in word frequency. \citet{cabanac-prevalence-2021} designed a paper generation detector based on syntax rules, which achieved 83.6\% detection precision in recognizing SCIGen-generated papers.

\subsection{Article type recognition}

Journals publish different types of articles, majorly including original research papers, review papers, commentaries, short reports, and letters. Some journals only accept original research papers, some only accept review papers and some publish multiple types of articles. Screening submissions for the right article types is basically a task of document classification. Conventional document classifiers are often based on traditional machine learning methods. Examples include Bayes classifier~\citep{nigam-text-2000} and Latent Dirichlet Allocation (LDA)~\citep{hingmire-document-2013}. As deep learning models are proved to deliver better performance with the use of semantic information, neural network-based methods have become the mainstream way for document classification. The publishing of recurrent convolutional neural networks~\citep{lai-recurrent-2015}, hierarchical attention networks~\citep{yang-hierarchical-2016}, Graph-CNN~\citep{peng-large-2018}, and BERT~\citep{adhikari-docbert-2019} has gradually improved the performance of document classification. These are all transferable frameworks and can be used for recognizing article types at the screening stage if corresponding labeled datasets are provided.

\subsection{Scope evaluation}

Every journal has its own aims and scope. At the screening stage, editors need to check the submissions to see whether they conform to the journals' objectives and whether the content is what the journal wants to deliver to its audience. The survey conducted by~\citet{froese-surviving-2019} reveals that a significant cause for desk-rejection is the mismatching with the journal's aims and scope. Tirthankar Ghosal and his teammates have done a large number of studies on the mismatching between submissions and the journal's aims and scope. They designed a binary classification model to help editors and authors to determine whether a manuscript matches with a journal~\citep{ghosal-is-2019}. They also used a multiview clustering method to identify mismatched submissions~\citep{ghosal-multiview-2019}. In 2020, they further found that titles and author profiles are more helpful in determining whether it is a good match between the manuscript and the journal. By doing so, mismatching of submissions and journals can be better identified~\citep{ghosal-empirical-2020}. In addition, the technique proposed for a related task of academic venue recommendation can also be employed here. \citet{wang-content-2018} was an early study into academic venue recommendation. In their study, abstracts of papers were crawled from web pages to train a chi-square feature selection and softmax regression hybrid model for recommendation. Content and Network-based Academic VEnue Recommender system (CNAVER)~\citep{pradhan-cnaver-2020} and Convolutional layer, LSTM with Attention mechanism-based scholarly VEnue Recommender system (CLAVER)~\citep{pradhan-claver-2021} are two recently proposed academic venue recommender systems. The former incorporated the paper-paper peer network model and the venue-venue peer network model, while the latter combined convolution, Bi-LSTM~\citep{graves-framewise-2005}, and attention mechanism~\citep{bahdanau-neural-2015}, both of which systems can both effectively solve the cold-start, sparsity, diversity, and stability problems.

\section{Main review}

In the traditional peer review process, after clearing the initial screening, submissions will be sent to external experts for review. Many publishers have developed reviewer guidelines for the references of the reviewers. In ASPR, the main review is conducted with computers taking up the workload from external experts instead. The main review stage is designed with a focus on the thorough reviewer guidelines provided by the top computer science conference of NeurIPS.\footnote{https://neurips.cc/Conferences/2022/ReviewerGuidelines} In addition to this, several other journal reviewer guidelines~\citep{song-scoping-2021} are also incorporated. References are also drawn on the studies on peer review~\citep{jefferson-measuring-2002,ernst-understanding-2021}. This approach ensures that the review process is informed by best practices in the field, while also drawing on insights from previous studies. Different aspects that should be focused on in the evaluation of the manuscripts are originality, soundness, clarity, and significance. In this section, we review the related technologies that can enable ASPR based on the four aspects.

\subsection{Originality}

Originality, or novelty is one of the most important criteria for scholarly publication. In the academic context, originality does not necessarily mean new inventions or discoveries. A study might not be able to reinvent the wheel in a certain field, but if it is a new idea that can move other ideas forward for an incremental amount of advance in current knowledge, it should be considered a study with originality. From this aspect, originality is defined as recombining the components of pre-existing knowledge in an unprecedented way~\citep{schumpeter-business-1939,nelson-evolutionary-1982}.

Based on this definition, reviewing a research paper for originality can be thought of as looking for the recombination of knowledge. Taking into consideration that the knowledge recombination is based on what can be found in the references section, the originality of a study can be evaluated based on the novel combinations of reference papers. \citet{dahlin-when-2005} proposed a definition of the radical invention and designed a measurement method for an invention's radicalness using backward citations for patents which can be also used for papers. In their method, novelty was measured by quantified citation similarity. A patent of radical invention should have a unique citation structure distinct from those of the existing patents. \citet{matsumoto-introducing-2021} later developed this measurement method towards a larger-scale novelty analysis. They only used bibliometric information for novelty measurement of papers in various research fields, countries, and time periods. \citet{uzzi-atypical-2013} assessed 17.9 million papers from the Web of Science to study the relations between the combination of a paper's references and its citation counts. Their findings suggest that the most influential science is based on highly typical combinations of earlier work, with the addition of combinations that have never been matched before. This can be the basis of distinguishing papers with originality from those without. \citet{shibayama-measuring-2021} designed a more integrated method, making use of both the citations and content of a paper. In their method, the semantic distance of references is quantified to determine the novelty of a given paper.

There are also methods to evaluate papers' originality from other perspectives. \citet{park-second-2014} introduced a method of novelty identification based on a generative model. In their method, the novelty of a paper is rated according to the likelihood of a paper being machine-generated. This is enabled based on the proposition that if a paper has great similarity with any machine generated paper, then it is less likely to be a novel research paper. \citet{amplayo-network-2018} presented a graph-based novelty detection model. In their model, authors, documents, keywords, topics, and words are used for feature representations to compose different graphs. Different papers added to the graphs will lead to different changes in the graphs. A paper that makes greater changes can be regarded as a paper with greater novelty.

\subsection{Soundness}

Soundness is a crucial aspect that distinguishes papers from general articles. The methods and argumentation presented in papers must be thorough and comprehensive, providing sufficient details to enable other researchers to reproduce the methods and replicate the results. Previous automated assessment of soundness has mainly focused on data quality, standard constraints, and references. Data quality is assessed primarily in terms of validity and accuracy. Standard constraints are assessed to determine whether the paper adheres to the requirements of relevant norms in scholarly publication. For references, the manuscript is scrutinized to ensure that all words and phrases that require citation are accurately cited.

Data accuracy and precision are crucial elements of high-quality papers. \citet{fanelli-how-2009} revealed that 1.97\% of researchers admitted to fabricating, falsifying, or modifying data or results at least once. The admission rate of data falsification in the survey of the colleagues' behavior stands at 14.12\%. These statistics highlight the presence of data falsification, which is a serious form of academic misconduct. Therefore, data validation is of utmost importance in the peer review process to ensure academic rigor. A common method for data validation in scientific research is the use of \textit{P} value. In academic disciplines, such as psychology and econometrics, a considerable amount of data are usually involved in the research. Validating data for this kind of research through the calculation of \textit{P} values is formidably laborious for human reviewers. To automate the calculation for data validation, \citet{nuijten-statcheck-2020} employed statcheck to identify statistical inconsistency. Statcheck can be used as an R package or in a browser to automatically extract statistical research results from PDF or HTML files and recompute the \textit{P} values to validate the data. Furthermore, statcheck's effectiveness in meta-analyses has also been explored.

In order to standardize the soundness of content in papers, various organizations have developed standardized reporting guidelines. Some famous standardized reporting guidelines are Consolidated Standards of Reporting Trials (CONSORT)~\citep{schulz-consort-2010}, STrengthening the Reporting of OBservational studies in Epidemiology (STROBE)~\citep{von-Strengthening-2007}, Standards for Reporting Diagnostic Accuracy (STARD)~\citep{bossuyt-stard-2015}, and Animal Research: Reporting of In Vivo Experiments (ARRIVE)~\citep{percie-arrive-2020}. These guidelines list in detail the information and processes that should be provided in a paper in the relevant field, with checklists for authors and reviewers to examine the specification and soundness of the paper. Automated checking of papers for compliance with these guidelines is key to ASPR in terms of soundness review, and StatReviewer is such a tool that not only checks the adherence to these guidelines, but also the accuracy of statistical methods used in the paper. This tool is now used in the Aries system for actual application in peer review.\footnote{\url{https://www.ariessys.com/newsletter/february-2018/new-decision-support-tool-statreviewer-available-in-15-0/}}

Proper referencing gives papers credibility and authority. Poor citation practices are a cause of concern to the authors' academic ability and can even become a suggestion of potential plagiarism. One poor citation practice is that authors fail to provide citations to relevant major studies. For the checking of referencing quality, techniques used in citation recommendation~\citep{ma-review-2020,ali-deep-2020} can be applied to detect poor citation practices. \citet{anderson-keep-2011} also proposed several methods to assess whether a cited paper has its content properly presented in the citing paper.

\subsection{Clarity}

Clarity in peer review is checked on the basis of linguistic quality, writing style, and text organization. Linguistic quality refers to the appropriate use of language, including error-free spelling, correct grammar, etc. Writing style is expected to use academic language and follow academic norms. Text organization requires that papers conform to publicly accepted writing structure.

The simplest way to check spelling errors is to have the word looked up in the speller lexicon, which can be enabled by algorithms like n-gram or Levenshtein distance. In this way, the detected spelling errors can also be corrected with the right version in the lexicon. Examples include the research conducted by~\citet{zamora-use-1981} and~\citet{hodge-comparison-2003}. However, these dictionary-based methods are slow and at the same time rely heavily on the size of dictionaries. In addition, they are not incapable of more complex checking of word types. For this reason, \citet{ahmad-learning-2005} tried to use the expectation maximization algorithm~\citep{dempster-maximum-1977} to learn the edit distance weight directly from the correction log of the search without checking the lexicon. For a more advanced deep learning method, \citet{whitelaw-using-2009} utilized big data in training to build seq2seq spelling error detection models to achieve more efficient detection and correction. Most of the tools for spelling detection are developed with artificially created datasets that lack context, so~\citet{flor-benchmark-2019} constructed an annotated dataset with contextual spelling errors from English language learners and designed a minimally supervised context-aware approach for spelling error correction.

Grammar check is the foundation task for grammar correction. It was the focus of both the CoNLL-2013~\citep{ng-conll-2013} and CoNLL-2014~\citep{ng-conll-2014} shared tasks on grammatical error correction. Like the way spelling check develops, the early methods in grammar check are basically rule-based and data-driven. The limitation of these methods is that they can only detect certain types of grammatical errors, like erroneous prepositions~\citep{chodorow-detection-2007,felice-automatic-2013} and erroneous determiners. In order to bypass the dependence on annotation data, \citet{liu-exploiting-2017} trained a neural network-based model for grammatical error detection using unlabeled generated data. Their research showed that generated errors were also effective for automatic grammatical error detection. Deep learning is also applied to grammatical error detection. \citet{rei-compositional-2016} treated grammatical error detection as a sequence marking task and used Bi-LSTM to predict the error tagging. \citet{bell-context-2019} made use of contextual word representations to represent words, learning semantic and component information in different contexts, and integrated the representation into the error detection model to improve the detection performance. \citet{wang-grammatical-2020} used the contextual information to represent words with BERT-based word embeddings. A synthetic grammar error corpus was also employed. They further designed a positive and negative sample pair training mechanism to capture differences for more effective grammatical error detection. \citet{hu-considering-2021} constructed a neural network-based correction model for grammatical errors. In their method, word vectors are used for feature representations instead of direct one-hot encoding to reduce semantic redundancy. In addition, they try to further compress article features for optimization using clustering methods.

Academic writing requires compliance with the formal and rigorous writing style, which is significantly different from other forms of writing. Papers written in an overly casual style may come across as unprofessional or questionable. An Automated Evaluation of Scientific Writing Shared Task was conducted in 2016~\citep{daudaravicius-automated-2015,daudaravicius-report-2016}. Its goal was to assess whether the style of a given sentence complied with the academic writing norms. For this shared task, the best performance was achieved by an attention-based encoder--decoder model developed by~\citet{schmaltz-sentence-2016}. The second best results were delivered by a convolutional neural network model~\citep{lee-ntnu-2016} that used both word2vec and GloVe~\citep{pennington-glove-2014} embedding methods. \citet{sanchez-combined-2016} employed Tree Kernel-SVM~\citep{cortes-support-1995,collins-new-2002} based methods and achieved the third best performance.

One common way to achieve clarity in academic writing is to structure the paper in a certain layout. There are studies making efforts to evaluate the organizational clarity of scientific papers based on their writing layout. For most scientific papers, the IMRaD is a standard structure, which includes four main sections: Introduction, Methods, Results, and Discussion. \citet{agarwal-automatically-2009} studied multiple approaches to the sentence-level classification of biomedical papers into those four IMRaD sections. In their study, an SVM classifier achieved the best results for automatic classification and annotation of sentences in biological papers.

\subsection{Significance}

As written in the Reviewer Guidelines of NeurIPS, when evaluating a paper for its significance, the reviewer should base on the following questions, ``Does the submission address a difficult task in a better way than previous work? Does it advance the state of the art in a demonstrable way? \ldots '' Significance of a paper can be understood as the impact it makes. From this perspective, evaluating the significance of a paper can be achieved by evaluating its impact, which to a great extent is reflected in its citation counts. A paper with major impacts naturally and actually earns more citation counts. Therefore, in ASPR, citation count prediction can be used to automate the evaluation of significance. Apart from this, the impact of some studies is also decided by the performance of their proposed methods. In this sense, significance evaluation can also be achieved by comparing the method performance with the benchmarks to see whether the method better than the SOTA methods or not.

Despite being a controversial measure for the impact of papers~\citep{brody-earlier-2006,wang-knowledge-2016}, citation count remains a popular method for this task. \citet{fu-models-2008} used titles, abstracts, keywords, and bibliometric features like the citation counts of the author's previous works and the author's academic backgrounds as feature representations. The task of citation count prediction is completed as a classification task using SVM. This method delivered outstanding results, but using bibliometric features for citation count predictions in ASPR can cause bias against researchers in their early careers and in favor of researchers from prestigious institutions. In addition to this, the main body of the research paper will be left out using this method, which will fail to deliver a thorough and all-around review of papers. The deep-learning model designed by~\citet{ma-deep-2021} is one of the pioneering studies to incorporate the overall content. It extracted semantic information from paper text first using doc2Vec~\citep{le-distributed-2014} and further using Bi-LSTM with attention mechanism for high-level features. This model achieved SOTA performance, outperforming the baseline models.

One of the important and more objective aspects of significance evaluation lies in the evaluation of research results. For scientific research papers that propose new methods, this can be achieved by SOTA comparisons, i.e.\ conducting performance comparisons with benchmark models. Authors often claim that their results reach the SOTA, but whether this is true or not needs to be verified, and it is possible that a better performing method exists that the authors are not aware of. Nevertheless, conducting SOTA comparisons to check their claims is no easy task. To enable SOTA comparisons in ASPR requires a thorough collection of all the existing benchmarks and this collection needs to be updated constantly to take in newly-made benchmarks. Papers With Code is a prestigious platform for gathering papers and source code in machine learning. This platform also collects benchmarks respectively for different tasks in different domains on its own and through crowdsourcing.\footnote{\url{https://paperswithcode.com/sota}} The benchmarks listed for separate tasks can be used for SOTA comparisons. \citet{hou-identification-2019} proposed a method to automatically compose datasets similar to the benchmarks of Papers With Code. This method, developing from natural language inference, learns the similarity patterns between labels and paper's content at word level to automatically extract tasks, datasets, evaluation metrics, and scores from published scientific papers. They then constructed a new corpus TDMSci composed of NLP papers manually annotated with the task, dataset, and metric at document level. Based on this corpus and using the triples of the task, dataset, and metric, they further designed a TDM tagger through data augmentation for automated extractions in the NLP domain~\citep{hou-tdmsci-2021}. \citet{kardas-axcell-2020} developed a machine learning tool AXCELL to extract results from research papers automatically. In AXCELL, the results extraction task is divided into three subtasks: table type classification, table semantic segmentation, and linking results to baselines in leaderboards. With these extraction methods, datasets of SOTA benchmarks in different domains can be built for systematic SOTA comparisons. Existing SOTA datasets can also be included to enrich these datasets. When ASPR receives a new submission, the system will extract the triple of core tasks, datasets, and metrics from the manuscript. This triple will then be compared to the triples of SOTA benchmarks to determine whether the method performance in this manuscript matches or exceeds the SOTA benchmarks in a certain domain. The significance of a study is reviewed based on this comparison.

\section{Review report generation}

A review report is the end product of the peer review process. Constructive review reports provide valuable feedback to authors for them to improve their manuscripts. A comprehensive review report should contain a summary of the manuscript, comments on the details, scores to quantify the evaluation, and the final editorial decision on accepting or rejecting the manuscript. Review report generation is of great importance in ASPR\@.

\subsection{Summarization}

According to most reviewer guidelines, a review report should first include a summary section, briefly outlining the manuscript and its main contributions. This summary presents the basic knowledge and understanding of the reviewer on the manuscript. In ASPR, this is a task of automated text summarization, a common task in NLP. \citet{mohamed-srl-2019} innovatively adopted a Wikipedia graph-based approach to build a summary generation model. In their model, explicit semantic analysis (ESA) is used to label words and represent them as vectors with weighted Wikipedia concepts. Semantic role labeling (SRL) is adopted to identify semantic arguments based on the predicate verbs in the sentence. The summarization is achieved by the construction of a concept graph representation on semantic role-based multi-node vertices. A neural abstractive summarization framework proposed by~\citet{pilault-on-2020} is capable of producing abstractive summaries for long documents. The hybrid framework is composed of a hierarchical encoder-based extractive model and a Transformer language model. An initial extractive step is undertaken to reduce the context for the next abstractive step. The input to the Transformer language model is reordered to identify the introduction, the extracted sentences, the abstract or summary, and the rest-of-the-article. This way the hybrid model is more focused on the summarization task and outperforms baseline methods on the arXiv, PubMed, and bigPatent datasets. \citet{gupta-effect-2021} conducted a study to improve model performance on the task of scientific article summarization and proved in experiments the advantages of transfer learning via intermediate pre-training for this task.

\subsection{Comment generation}

The second section of a review report should present detailed review comments on every aspect of the manuscript. Automatic comment generation for papers is a challenging task for ASPR. There exists already relevant research on this field, but it remains inadequate. Many existing studies have adopted an NLP approach in the generation of review comments for papers using slot filling models. In NLP-enabled slot filling, the models generate review comments by filling in a preset review format with extracted information. \citet{bartoli-your-2016} proposed a corpus-based method to generate review comments based on the full text of a paper and a preset overall evaluation. \citet{wang-reviewrobot-2020} designed a knowledge-driven end-to-end framework ReviewRobot to automatically generate knowledgeable and explainable scores and comments. They achieved this by comparing the knowledge graphs built from the given paper and a large number of other papers in the same domain. \citet{yuan-can-2022} built a paper dataset and annotated the review comments for different aspects, so as to train a review comment generation model by using BART~\citep{lewis-bart-2020}. Experiments show that their model is capable of generating comprehensive comments, but they considered that more improvements were needed for this task.

In spite of the deficiency in the research into review comment generation for papers, some efforts have been made in relevant tasks, including customer review generation and news comment generation, etc. With proper training corpora, neural network-based methods in these studies can be applied to generate review comments for papers. \citet{baroni-linguistic-2019} proved in experiments that syntactic rules can be captured through deep learning and further used to generate meaningful natural-language sentences. Apart from syntactic generation methods, topic, and semantics are also used for review generation. \citet{li-generating-2019} proposed an aspect-aware generation model and made full use of semantic, syntactic, and contextual information. In their aspect-aware model, the reviewing of each aspect is set as a main task and this main task is assisted by auxiliary tasks. Two decoders are used in their model with one predicting a structural draft and another filling in words. Through this aspect-aware coarse-to-fine generation process, the model delivers a great performance in review generation in terms of overall generation quality, aspect coverage, and fluency.

\subsection{Scoring}

Scores to quantify the review comments are generally included in the review report. This is particularly common in peer review of conference papers. In ASPR, automated paper scoring is also a crucial stage that can only be completed based on all the results from the previous stages. There are relevant techniques that can be used to fulfill this task.

The task of automated paper scoring is considered by most researchers as a multiclass classification problem, i.e.,~treating each score as a class and solving it using classification techniques. \citet{qiao-modularized-2018} designed an attention-based modularized recurrent convolutional neural network to produce scoring on various aspects of papers, including appropriateness, clarity, originality, etc. Experiments showed that their method outperformed two baseline methods on the average quadratic weighted kappa. In addition to scoring different aspects of a paper, there are also methods proposed to produce an overall score directly. \citet{leng-deepreviewer-2019} introduced an attention-based framework DeepReviewer that assigns scores for papers on OpenReview based on the semantic, grammatical, and innovative features combined. This framework is composed of a hierarchical recurrent convolutional neural network, a customized unsupervised deep context grammar model, an unsupervised high-dimensional spatial density-based innovation model, and an attention layer to generate the final review score. Experimental results showed that DeepReviewer outperformed many baseline models. \citet{li-multi-2020} presented a multi-tasking shared structure encoding method that can choose shared network structures and auxiliary resources in an automatic way. Their method is especially helpful in the case of insufficient data.

\subsection{Decision making}

The prediction task of review decision on the acceptance or rejection of a manuscript can be recognized as a binary classification problem. \citet{yang-automatic-2018} proposed a model of modularized hierarchical convolutional neural network to predict the acceptance result. This model was trained on positive samples of published arXiv preprints and negative samples of unpublished arXiv preprints. Experimental results showed that the model achieved 67.7\% accuracy in its prediction. Additionally, in their study, the influence of authors, abstract, conclusion, and title on the prediction were also analyzed and authors were found to have a greater impact on the results. \citet{skorikov-machine-2020} built a machine learning-based model to predict paper acceptance in prestigious AI conferences. In their study, a comparison was made between seven different machine learning algorithms. Random forest~\citep{breiman-random-2001}, a classifier consisting of many decision trees delivered the best results. The model that used this classifier achieved an accuracy of 81\% on the PeerRead dataset. \citet{lamarre-textual-2021} analyzed the full text of both accepted and rejected manuscripts to explore their semantic, lexical, and psycholinguistic feature differences. They found that the readability of accepted manuscripts was lower than that of rejected manuscripts. By using a logistic regression of bag-of-words to predict the peer review outcome, they found that their model performed best when using the introduction text for prediction. \citet{bao-predicting-2021} proposed an algorithm to build up decision sets for acceptance prediction of papers in a simple, effective, and interpretable way.

Apart from the textual content of a paper, its structure and layout also make a difference in its quality~\citep{sun-structuring-2014}. \citet{huang-deep-2018} treated reviewing paper as an image classification task and trained a classifier that built on deep convolutional neural networks to predict the acceptance results for papers based solely on their visual features. He further provided tools that directly learned the mapping in the image space to provide authors with suggestions to enhance their papers visually. Moreover, visual features were also combined with text features to conduct document quality evaluation in the study of~\citet{shen-joint-2019}. They used Inception V3 to generate visual feature embedding of a manuscript's snapshot and Bi-LSTM to produce textual feature embedding. The two embeddings were used to train a classification model that delivered the SOTA performance on the PeerRead dataset.

\section{Current challenges}

In the sections above, we explore the existing technologies for achieving ASPR. Through the review, relevant implementation is found available at each stage of ASPR. Nonetheless, ASPR, as a highly demanding task at its early phase, is in the face of many challenges. In this section, we discuss some of the major challenges.

\begin{itemize}

\item \textbf{Inadequate data}

Through our review, we can see that there are indeed quite a few datasets available for ASPR. But it is still far from being enough and data insufficiency is still a major problem in ASPR. For one thing, existing datasets do not cover all the fields in academic studies. The vast majority of them only focus on computer science, since it is the field that is the most closely related. The ideal dataset to achieve the best performance in ASPR within the interdisciplinary trend should be a complete collection of all the papers from all different domains. Moreover, these data need to be structured in a predefined manner for machine learning. There are four main review dimensions in ASPR, which are originality, soundness, clarity, and significance. Each dimension shall be treated as a sub-task with corresponding datasets provided. This means that the review comments need to be labeled and segmented based on the content so as to build these corresponding datasets for the four review dimensions. Most of the available review comments, like those on OpenReview, are usually in plain text without content labels, therefore efforts are needed to create separate sub-datasets for each review dimension.

To add to the problem of data insufficiency is data imbalance~\citep{santos-unifying-2023} between accepted manuscripts and rejected manuscripts. Existing datasets are mostly composed of peer-reviewed published papers. With these datasets, computers can learn to recognize papers of good quality. However, for those papers with insufficient quality, computers lack enough learning materials to identify what are bad papers. Some platforms, OpenReview for example, provide the public access to rejected papers and the review comments on them, but most rejected papers are not made public, especially those papers that are desk-rejected. Some datasets like PeerRead do claim to include rejected papers. But it is hard to be sure whether these papers are really peer-reviewed and rejected as it is not confirmed officially by the academic publishers.

``If you build a decision-making system based on the articles which your journal has accepted in the past, it will have in-built biases''.~\citep{heaven-ai-2018} Data imbalance gives rise to even more severe big data bias. The current ASPR is largely enabled by deep learning and big data, which usually means building models that are trained on certain datasets. Thus, the data distribution and data features of these datasets affect the learned models. For one thing, computers might not be able to identify papers with great novelty as these papers are in minority and deviate from the learned patterns of quality papers. Human reviewers with knowledge and experience might recognize these papers through acute senses, but computers will tend to reject such papers as they are off the beaten track. For another, computers trained with imbalanced data are more capable of recognizing quality papers and are less sensitive to detect flaws in manuscripts. This can lead to inappropriate acceptance of unqualified papers.

\item \textbf{Imperfect document parsing and representation}

As reviewed in Section~\ref{sec:parsing-representation}, there are already related techniques available for document parsing and representation to achieve ASPR, but necessary improvements are still needed to be made. First of all, the accuracy in the parsing of PDF files needs to be improved. Currently, existing parsers are highly capable of extracting content and structure from PDF files, but they still fall short in the face of some special characters and unusual branches of typography. To live up to the rigor of reviewing, parsers used in ASPR should not be allowed to make even one punctuation error. However, this is still not realized at the very moment. A compromise solution will be requiring the writing on LaTeX and the submission of all related LaTeX files.

Second, the representation of long documents needs to be refined. Papers are generally of great length. However, with the current presentation methods, long documents can only be poorly represented because of their length. Available long document representation methods demand powerful hardware and are computationally ineffective. This task is greatly in need of further study.

Last but not least, parsing of other types of resources, like videos, websites, and source code, also needs to be developed. With the advancement of technology, the content of academic writing is extended to include more supplemental materials, such as demonstration videos, related websites, and corresponding source code. To cope with the new changes, multimodal parsing has become a hot spot in recent studies~\citep{zhang-multimodal-2020,uppal-multimodal-2022}. In ASPR, computers should be able to review all different types of data, which can only be enabled by multimodal parsing technology.

\item \textbf{Defective human--computer interaction}

The interactions between reviewers and authors are at the core of the whole peer review process. A manuscript might be accepted, revised, or rejected in traditional peer review. These decisions are made by the editors majorly based on the review comments and also in some cases the interactions between the reviewers and the authors. These interactions are especially important for those manuscripts that are revisions. In the case of those manuscripts that are revisions, usually after the first round of peer review, authors of these manuscripts will be provided with feedback from the reviewers, based on which they can make proper revisions to improve the manuscripts. In this review process, both the reviewers and the authors should communicate with each other in order to properly address all concerns about the manuscript. These interactions provide the editor with important information for the final editorial decision on whether a manuscript should be accepted for publication or not. In ASPR, technologies required for interactions between the ASPR system and the authors are still not mature enough. Therefore, in the early phase of ASPR that we propose in this paper, interactions between the computers and the authors are not included. If a manuscript is rejected by the ASPR system, after making revisions based on the computer-generated comments, the author can submit the revised manuscript to ASPR for reviewing as a new manuscript.

\item \textbf{Flawed deep logical reasoning}

Peer review is a highly demanding task for reviewers' reasoning ability. Reviewers need to read through the whole manuscript to scrutinize the consistency and soundness in the study and the writing. Some examples of the issues that need to be assessed by reviewers with strong reasoning ability include: are the methods used able to answer the research questions; do the conclusions match the research results and the research aims? \ldots To answer these questions, reviewers are required to be knowledgeable in a certain field and capable of logical reasoning. Induction, abduction, and deduction are all closely involved in the peer review process, making it an error-prone process. Errors here mean that those logical flaws are not detected in time and thus the problematic manuscript is accepted for publication. For human reviewers, verifying a study's consistency and soundness through logical reasoning is no easy task. For computers, it is even more so. To achieve this part in ASPR relies on the full realization of automated reasoning. There are related studies~\citep{antoniou-survey-2018,chen-review-2020,storks-recent-2020}, but there is still a certain period of development to be used for ASPR. Besides, academic studies are trending towards an interdisciplinary future. The integration of knowledge from multiple fields also poses greater difficulty to the knowledge-based logical reasoning of computers.

\end{itemize}

In summary, the major challenges hindering the full realization of ASPR are also major issues in certain subfields in AI. From this, we can see that ASPR is an advanced integration of AI technologies. Its realization and the new era of strong AI will come hand in hand.

\section{Future directions}

The development of ASPR can be divided into two phases: the coexistence phase and the independent phase. The difference between these two phases lies in whether ASPR is mature enough to generate completely explainable, reasonable, and comprehensive review reports. ASPR in its early phase will coexist with peer review; in other words, ASPR is serving as an assistant for humans in its early phase. Currently, ASPR is in this coexistence phase. As has been established from our review above, there are practical technologies to achieve ASPR. Nonetheless, some challenges and issues are yet to be resolved. In the following of this section, we explore major future efforts that are necessary for developing ASPR into its independent phase.

Firstly, construct the greatest possible amount of large-scale labeled datasets~\citep{paullada-data-2021}. These datasets should cover every review dimension of ASPR. Data labeling is a labor-intensive and time-consuming task to be accomplished by the cooperation of the entire community. ASPR research will always go hand in hand with the building of labeled datasets, which are the foundation of ASPR. In particular, efforts should be made to collect rejected manuscripts in a reasonable way. One possible solution is to encourage the inclusion of the titles of the submitted journals as the required metadata for manuscripts submitted to preprint servers.

Secondly, build a continuously updated knowledge graph~\citep{zhong-comprehensive-2023} to support fact check in ASPR. This knowledge graph is also the foundation to achieve fully independent ASPR. In terms of the ultimate goal, the knowledge graph should include all knowledge existing in the world, or at least it should cover all knowledge related to the certain fields that use ASPR. Moreover, timely updating of the knowledge graph is needed to keep track of fast-paced knowledge renewal.

Thirdly, embrace the changes brought by ChatGPT\footnote{\url{https://openai.com/blog/chatgpt/}} and other large language models (LLMs)~\citep{zhao-survey-2023}. As a chatbot with LLM at its core, or what some researchers consider to be a comprehensive application tool of AI, ChatGPT has not only been greatly discussed in the AI research field, but has also brought enormous impact on a wide range of industries. ASPR research should actively react to the research paradigm shift and other innovations that come with ChatGPT in order to better advance related technologies.

Fourthly, strengthen the study on the automated evaluation of papers' core values such as originality and significance. These values are the spirit of papers. As reviewed and discussed above, some researchers have proposed related methods to automatically evaluate these core values of papers, but more efforts are still needed. Automated evaluation of papers' core values is the key point of achieving the independent phase of ASPR, thus primary research efforts shall be devoted to this area.

Fifthly, enhance the explainability of AI~\citep{vilone-notions-2021} in ASPR. Currently, most of the methods implemented in ASPR are driven by deep learning. This can cause the explainability issue to the outcomes of ASPR. Repeatability, verifiability, and explainability are indispensable for sound paper reviewing. To achieve explainability, the reasoning behind the review reports needs to be presented together in the outcomes. This can help ASPR to be a more reliable and convincing technology.

Sixthly, implement a unified end-to-end ASPR model for review report generation. A pipeline model is adopted as the present ASPR system corresponding to each step in the peer review process. The pipeline model is easy to implement, but errors tend to accumulate in the process and overall optimization is harder to be conducted. For the development of ASPR, an end-to-end model that conducts all reviewing steps simultaneously is more desirable. Review comments generated by end-to-end ASPR will be more comprehensive and thoughtful as different dimensions are reviewed inclusively. The design will be much more demanding, but it is well worth in-depth research.

With more devoted studies and the arrival of strong AI, it is merely a question of time before ASPR is able to develop into the independent phase. This process is estimated to come in the 2050s.

\section{Discussion of morality and ethics}

The rapid development of AI has been accompanied by increasing discussions about its morality and ethics~\citep{jobin-global-2019,hagendorff-ethics-2020}. ASPR, which is an integrated application of AI, faces the same situation. In this section, we discuss the possible moral and ethical issues of using ASPR in terms of the different phases of its development.

In the coexistence phase of ASPR and peer review, ASPR is acted as an assistant role, and humans are who give comments and make final decisions. So, it is also humans who bear the responsibility in the peer review. The main concern at this phase is the proper use of ASPR. As editors, while ASPR is making the screening results, especially the decision of desk-rejection, they must also read the manuscript personally to determine whether the manuscript is really not necessary to send for external review and avoid rejecting valuable manuscripts. As reviewers, they are required to review the manuscript carefully and write their review reports independently. On this basis, they should compare the review reports given by ASPR with their owns, judge whether their comments are omitted or biased, and improve their review reports accordingly. The ASPR review reports could not be adopted directly. In general, in the coexistence phase, humans cannot use the results of ASPR to replace their own work.

In the independent phase of ASPR, ASPR no longer requires human participants, and ASPR itself is the main body of responsibility. Some regular and often discussed moral and ethical issues of AI also appear on ASPR during this phase. The first one is authority. ASPR, in the independent phase, combines the roles of editors and reviewers in traditional peer review and has enormous power. Where does the authority of this power come from? Who is responsible for and supports such authority? By convention, before the launch of such an AI system like ASPR, an evaluation should be conducted by a qualified independent third-party organization. The evaluation should involve the algorithm, training data, and performance of ASPR. In this way, a series of questions deserve to be discussed. What are the criteria for ASPR to be able to enter the independent phase? Can it be bounded by meeting or even exceeding the performance of human experts? If yes, how should the standard be developed? Which human experts should be selected as representatives to benchmark this standard? These are questions highly difficult to reach an agreement on but well worth exploring. The second one is the issue of responsibility. If ASPR makes mistakes, such as giving insulting and discriminatory comments, providing incorrect revision suggestions, rejecting valuable manuscripts, and admitting defective manuscripts, whether relevant people and organizations are jointly and severally responsible? Who are they? Should it be the users, the developers, or the third-party appraisal organization that grants approval for the usage of ASPR? Generally speaking, the main issues to consider in the independent phase are authority and responsibility.

Moreover, whether in the coexistence phase or in the independent phase, authors should not deliberately write manuscripts in unconventional ways in order to obtain positive feedback in ASPR. Still less should they use the adversarial attack~\citep{zhang-adversarial-2020,ren-adversarial-2020} to train and generate manuscripts that can obtain high scores. This completely deviates from the original purpose of ASPR, and also makes academic writing lose its purity. ASPR providers should not use the manuscripts submitted by authors for other purposes, strictly enforce the confidentiality agreement in peer review, and use encryption to save the manuscripts and other metadata uploaded by users.

\section{Conclusion}

In this paper, we propose the concept of automated scholarly paper review (ASPR) and enunciate the review subject, review object, review scenario, output, and core term in the concept. We divide the pipeline of ASPR into four main stages: parsing and representation, screening, main review, and review report generation. We review the relevant literature and technologies as a means of answering the question that whether AI is already developed well enough to review scholarly papers independently. Through our review, we found that there are already corresponding technologies and resources that have been applied or can be applied to each stage of ASPR for its implementation. Taking into consideration the surging number of manuscripts and the steady development of relevant studies, ASPR is evaluated to be of great potential and feasibility. We further discuss the current challenges hindering the full realization of ASPR. The major challenges in its development are inadequate data, imperfect document parsing and representation, defective human--computer interaction, and flawed deep logical reasoning. We explore the future directions of ASPR and discuss its potential moral and ethical concerns. In summary, the application and development of ASPR will bring academic publishing into a fairer, more efficient, and more scientific era. In its full realization, computers will be able to carry the onerous reviewing workload for humans and both the research and publishing circles will benefit greatly from ASPR. Before this can come true, ASPR will continue to coexist with peer review as reinforcement in academic publishing.

\section*{Acknowledgments}
This work is partly funded by the 13th Five-Year Plan project Artificial Intelligence and Language of State Language Commission of China (Grant No. WT135-38). Grateful appreciation is given to the reviewers for their insightful comments and valuable suggestions. Special and heartfelt gratitude goes to the first author's wife Fenmei Zhou, for her understanding and love. Her unwavering support and continuous encouragement enable this research to be possible.

\section*{Compliance with ethical standards}
\textbf{Conflict of interest} The authors declare that there is no conflict of interest regarding the publication of this paper.


\bibliography{sn-article}% common bib file


\end{document}